\documentclass{article}

\PassOptionsToPackage{numbers,sort&compress}{natbib}
\usepackage[preprint]{neurips_2026}

\usepackage[utf8]{inputenc}
\usepackage[T1]{fontenc}

\usepackage{hyperref}
\usepackage{url}
\usepackage{booktabs}

\usepackage{amsfonts}
\usepackage{amsmath}
\usepackage{amssymb}
\usepackage{mathtools}
\usepackage{amsthm}        % ADDED: needed for proposition/lemma/remark environments.

\usepackage{nicefrac}
\usepackage{microtype}
\usepackage{xcolor}

\usepackage{graphicx}
\usepackage{subcaption}
\usepackage{wrapfig}
\usepackage{multirow}
\usepackage{array}
\usepackage{enumitem}

\usepackage{tikz}
\usetikzlibrary{positioning,fit,calc,arrows.meta}

\usepackage[most]{tcolorbox}
\tcbuselibrary{skins,breakable}

\usepackage{algorithm}
\usepackage{algpseudocode} % CHANGED: use algpseudocode instead of algorithmic.

\setlist[itemize]{leftmargin=*, itemsep=2pt, topsep=2pt}
\setlist[enumerate]{leftmargin=*, itemsep=2pt, topsep=2pt}

\allowdisplaybreaks      % ADDED: useful for appendix proofs.

\newtheorem{proposition}{Proposition}

\title{Path-Space Mirror Descent for On-Policy Reinforcement Learning under the Generalized Schrödinger Bridge}

\author{
\parbox{\textwidth}{
\centering
{\small
Yuehu Gong\textsuperscript{1}
\quad
Zeyuan Wang\textsuperscript{2}
\quad
Yulin Chen\textsuperscript{2}
\quad
Shutong Ding\textsuperscript{3}
\quad
Qingyuan Zhou\textsuperscript{4}
\quad
Yanwei Fu\textsuperscript{1,5}
}
\\[0.6em]
{\small
\textsuperscript{1}School of Data Science, Fudan University
\\
\textsuperscript{2}Laboratory for Big Data and Decision, National University of Defense Technology
\\
\textsuperscript{3}ShanghaiTech University
\\
\textsuperscript{4}College of Computer Science and Artificial Intelligence, Fudan University
\\
\textsuperscript{5}Shanghai Innovation Institute
}
}
}

\begin{document}

\maketitle

%%%%%%%%%%%%%%%%%%%%%%%%%%%%%%%%%%%%%%%%%%%%%%%%%%%%%%%%%%%%
% Main paper
%%%%%%%%%%%%%%%%%%%%%%%%%%%%%%%%%%%%%%%%%%%%%%%%%%%%%%%%%%%%

\begin{abstract}
Classical on-policy algorithms such as PPO and mirror descent policy optimization provide stable proximal policy updates through tractable action likelihoods, but are typically instantiated with simple Gaussian policies whose expressiveness can be limited in complex continuous-control tasks.
Generative policies based on diffusion and flow models provide more expressive action distributions, but they naturally define distributions over multi-step denoising paths whose terminal action density is often intractable, creating a mismatch with likelihood-based on-policy proximal updates.
To address this mismatch, we introduce \textbf{GSB-MDPO} (\emph{Generalized Schr\"odinger Bridge Mirror Descent Policy Optimization}), which formulates on-policy generative policy optimization as a Generalized Schr\"odinger Bridge problem over state-conditioned generation paths and instantiates the resulting path-measure update through mirror descent policy optimization.
The key insight is that the GSB path-space KL plays the role of the proximal term in MDPO while upper-bounding the terminal action KL, enabling direct control of the executed action distribution without explicit terminal action likelihood evaluation.
Experiments on 14 continuous-control tasks across Playground and Gym-MuJoCo demonstrate the empirical effectiveness of GSB-MDPO and support path-space regularization as a principled proximal update for multi-step generative policies.
\end{abstract}

\section{Introduction}
\label{sec:introduction}
% 引出我们文章的setting
On-policy reinforcement learning is widely used in continuous control due to its highly parallelizable rollout collection~\citep{mnih2016asynchronous, makoviychuk2021isaac}. Methods such as TRPO and PPO stabilize these updates by constraining the policy update through trust regions, likelihood ratios, or KL regularization~\citep{schulman2015trpo,schulman2017ppo,tomar2020mdpo}.
% 讨论guassian policy vs generative policy
Most implementations, however, rely on computationally efficient but restrictive policy classes such as diagonal Gaussians, which can limit performance when effective control requires multimodal action distributions.
Recently, generative policies based on diffusion~\citep{ho2020ddpm,song2021sde} or flow models~\citep{lipman2023flowmatching,geng2025meanflow} provide a more expressive alternative.
Generative policies have shown promise in planning~\citep{janner2022diffuser}, imitation learning~\citep{chi2023diffusionpolicy}, offline RL~\citep{wang2022diffusionql, wang2026one, park2025fql}, and online control~\citep{hansen2023idql,kang2023edp}.

% 引出generative policy的问题
Unlike Gaussian policies with tractable action densities, generative policies generate actions as the endpoints of multi-step denoising or transport processes. The resulting terminal action likelihood is often intractable or costly to evaluate~\citep{chen2018neuralode,grathwohl2019ffjord,zhang2025reinflow,zhang2025sacflow}, making standard on-policy updates based on likelihood ratios or action-space KL regularization difficult to apply.
% 现有的一些解决方案
Recent work has sought to make on-policy optimization compatible with generative policies by restoring or approximating the likelihood-ratio terms used in standard proximal updates. DPPO adapts policy-gradient updates to diffusion policies~\citep{ren2024dppo,liu2026diffusion_dcppo}, GenPO introduces invertible diffusion policies to recover exact action likelihoods~\citep{ding2025genpo}, and FPO constructs a surrogate policy ratio from a conditional flow-matching objective~\citep{mcallister2025fpo}.

% 引出关键问题
These methods make on-policy generative RL increasingly practical, but leave open a more fundamental question:
\emph{when terminal action likelihoods are intractable, where should proximal regularization be imposed for generative policies?}

% 引出我们对这些方案的认识
A suitable regularizer should be tractable while still controlling the distribution of executed actions.
The terminal action density often fails this requirement, whereas the full generation path provides a tractable alternative. Specifically, a generative policy induces a \emph{path measure} over its denoising trajectory, and the KL between path measures upper-bounds the KL between their terminal action marginals.
Thus, regularizing the current path measure toward the old one controls the executed policy without explicitly evaluating terminal action likelihoods.
This path-space perspective is aligned with generalized Schr\"odinger bridge formulations, which optimize stochastic path measures relative to reference processes~\citep{leonard2013schrodinger,liu2024gsbm}.

This leads to \emph{GSB-MDPO}, a path-space mirror descent framework for on-policy generative reinforcement learning. At each iteration, GSB-MDPO improves the old-policy advantage of the terminal action while regularizing the current generation path measure toward the old one through a path-space KL.
Both the advantage term and the forward path-KL term can be estimated from old-policy rollout paths via importance sampling, so policy updates do not require additional sampling from the current policy. For stochastic generators with shared diffusion coefficients, the same path-KL further reduces to a drift-MSE regularizer through the standard SDE path-KL identity.

% Empirically, We evaluate GSB-MDPO on two continuous-control benchmark suites, Playground and
% Gym-MuJoCo, together with ablation studies and toy analyses. Across these
% settings, GSB-MDPO provides a stable and effective path-space alternative to
% likelihood-ratio-based PPO-style updates for generative policies. 
Overall, our contributions are summarized as follows:
\begin{itemize}
    \item \textbf{GSB-MDPO: path-space mirror descent for generative policies.}
    We introduce \emph{GSB-MDPO}, a mirror-descent policy optimization framework
    that lifts the proximal update from terminal action distributions to
    conditional generation path measures under the generalized Schr\"odinger
    bridge perspective.

    \item \textbf{Terminal-action control via a path-KL upper bound.}
    We show that the KL between full generative path measures upper-bounds the KL between terminal action distributions.
    This allows GSB-MDPO to constrain the executed action distribution through a tractable path-space regularizer, without explicitly evaluating the terminal action likelihood.

    \item \textbf{Empirical validation on continuous-control benchmarks.}
    We evaluate GSB-MDPO on Playground and Gym-MuJoCo against representative
    Gaussian and generative on-policy baselines, including PPO, DPPO, FPO, and
    GenPO. Ablations on path regularization, reference mixing, ratio
    stabilization, and deterministic evaluation further support the role of
    path-space regularization.
\end{itemize}

\section{Related Work}
\label{sec:related_work}

\subsection{Generative Policies for On-Policy Reinforcement Learning}
\label{subsec:rw_onpolicy_generative}

Recent work has begun to integrate expressive diffusion and flow policies into explicitly on-policy reinforcement learning.
DPPO applies policy-gradient fine-tuning to diffusion policies and treats the denoising process as part of the policy optimization procedure~\citep{ren2024dppo}.
GenPO uses dummy actions to make diffusion policies invertible and likelihood-tractable for PPO-style updates~\citep{dinh2014nice,dinh2016realnvp,kingma2018glow,ding2025genpo}. CPPO applies proximal updates to conditional Gaussian denoising transitions, avoiding full denoising likelihoods but benefiting less in unimodal settings~\citep{liu2026diffusion_dcppo}.
FPO instead avoids exact likelihood computation by using a conditional flow-matching loss to construct a surrogate advantage-weighted ratio for flow policies~\citep{mcallister2025fpo}.
ReinFlow injects learnable noise into flow matching policies so that the flow becomes a discrete-time Markov process with tractable likelihoods for online RL fine-tuning~\citep{zhang2025reinflow}, while Flow-GRPO studies online policy-gradient training for flow matching models in the GRPO setting~\citep{liu2025flowgrpo}.
FlowRL further systematizes this area through a modular JAX framework and taxonomy for diffusion/flow-policy RL across Gym, DMC, and IsaacLab benchmarks~\citep{gao2026flowrl}.

GSB-MDPO differs from these approaches in both objective and implementation.
Rather than modifying the generative process to recover exact terminal action likelihoods, as in GenPO and ReinFlow, or replacing the likelihood ratio with a flow-matching surrogate, as in FPO, we formulate the update directly as mirror descent over path measures.
Both the advantage term and the forward path-KL term are evaluated on old-policy rollout paths through importance sampling.
This yields an on-policy path-space update motivated by the generalized Schr\"odinger bridge view, while preserving the data-reuse structure of PPO-style training.

\subsection{Other Generative Reinforcement Learning Paradigms}
\label{subsec:rw_other_generative_rl}

A large body of work studies generative policies outside the on-policy setting.
In imitation learning and planning, diffusion models have been used to generate trajectories or action sequences from demonstrations and offline data~\citep{janner2022diffuser,chi2023diffusionpolicy}.
In offline RL, Diffusion-QL introduced diffusion policies as expressive policy classes regularized by Q-learning objectives~\citep{wang2022diffusionql}.
Subsequent work improved the efficiency or compatibility of diffusion policies with offline RL objectives, including IDQL\citep{hansen2023idql}, EDP\citep{kang2023edp}, and flow-based Q-learning~\citep{park2025fql,fql_shortcut,mu2026deflow}.
These methods highlight the representational benefits of generative policies, but they learn from fixed datasets rather than fresh on-policy rollouts.

Generative policies have also been explored in online and off-policy actor-critic settings.
QSM uses Q-score matching to guide diffusion policies with value gradients~\citep{psenka2023qsm}, while QVPO derives a Q-weighted variational policy optimization objective for diffusion-based online RL~\citep{ding2024qvpo}.
DACER extends diffusion policies to online actor-critic learning with entropy regularization~\citep{wang2024dacer}, and DACERv2 improves diffusion efficiency by adding Q-gradient field supervision and temporal weighting to achieve strong performance with fewer diffusion steps~\citep{wang2025dacerv2}.
DIME derives a maximum-entropy objective for diffusion policies via marginal-entropy lower bounds~\citep{celik2025dime}.
Related flow-based approaches such as SAC Flow~\citep{zhang2025sac} make entropy maximization tractable
through velocity reparameterization. FLAC further
connects maximum-entropy generative RL with generalized Schr\"odinger
bridges and kinetic-energy regularization~\citep{lv2026flac}.
These works are complementary to ours: they mainly study offline, off-policy, or maximum-entropy actor-critic paradigms, whereas GSB-MDPO focuses on on-policy mirror descent over old-policy generative paths. 

\section{Preliminaries}
\label{sec:preliminaries}

\subsection{On-Policy Mirror Descent Policy Optimization}
\label{subsec:prelim_mdpo}

We consider a discounted Markov decision process
$\mathcal M=(\mathcal S,\mathcal A,p,r,\rho_0,\gamma)$.
A stochastic policy $\pi_\theta(a\mid s)$ is optimized to maximize
\begin{equation}
    J(\pi_\theta)
    =
    \mathbb E_{\pi_\theta}
    \left[
        \sum_{h=0}^{\infty}
        \gamma^h r(s_h,a_h)
    \right],
    \label{eq:rl_objective}
\end{equation}
where $h$ indexes environment interaction time.
For policy $\pi$, let
\begin{equation}
    A^\pi(s,a)=Q^\pi(s,a)-V^\pi(s)
\end{equation}
be the advantage function.
At iteration $k$, on-policy methods collect trajectories from the old policy $\pi_k$ and estimate
$A_k(s,a)\approx A^{\pi_k}(s,a)$, often using generalized advantage estimation~\citep{schulman2015gae}.
Let $d_k$ denote the state visitation distribution induced by $\pi_k$.

Mirror descent policy optimization formulates policy improvement as repeatedly solving a regularized local update around the old policy~\citep{geist2019regularized,tomar2020mdpo}.
In its ideal action-space form, the update is
\begin{equation}
    \pi_{k+1}
    =
    \arg\max_{\pi}
    \mathbb E_{s\sim d_k}
    \left[
        \mathbb E_{a\sim \pi(\cdot\mid s)}
        [A_k(s,a)]
        -
        \alpha
        D_{\mathrm{KL}}
        \left(
            \pi(\cdot\mid s)
            \|
            \pi_k(\cdot\mid s)
        \right)
    \right],
    \label{eq:action_mdpo}
\end{equation}
where $\alpha>0$ controls the proximity to the old policy.
For fixed $s$, the exact nonparametric optimizer is the exponential tilt
\begin{equation}
    \pi^*(a\mid s)
    =
    \frac{1}{Z(s)}
    \pi_k(a\mid s)
    \exp
    \left(
        \frac{A_k(s,a)}{\alpha}
    \right),
    \label{eq:action_mdpo_solution}
\end{equation}
with $Z(s)$ the normalizing constant.
In parametric policy classes, MDPO approximately optimizes this KL-regularized objective by gradient updates, retaining the mirror-descent principle while avoiding an exact trust-region solve~\citep{tomar2020mdpo}.
GSB-MDPO lifts this action-space update from the terminal action distribution to the full conditional path measure induced by a multi-step generative policy.

\subsection{Generalized Schr\"odinger Bridge for Path-Space Policies}
\label{subsec:prelim_path_gsb}

Iterative generative policies define action distributions through stochastic transport processes rather than one-step action densities~\citep{ho2020ddpm,song2021sde,lipman2023flowmatching}.
Let $t\in[0,1]$ denote generation time.
We model action generation by the SDE
\begin{equation}
    da_t
    =
    f_\theta(a_t,t,s)\,dt
    +
    \sigma_t\,dW_t,
    \qquad
    a_0\sim \mu_{\mathrm{init}},
    \label{eq:prelim_sde_policy}
\end{equation}
where $f_\theta$ is a learned drift or velocity field, $\sigma_t$ is a prescribed noise scale, and $W_t$ is a standard Wiener process.
The initial variable $a_0$ is sampled from a simple prior $\mu_{\mathrm{init}}$, and the terminal variable $a_1$ is the action executed in the environment.
Thus, a generative policy induces a path measure over trajectories from the prior endpoint to the action endpoint.

The Generalized Schr\"odinger Bridge (GSB) formulates stochastic generation as utility-regularized path-measure optimization relative to a reference process~\citep{benamou2000computational,leonard2013schrodinger,pavon2021data,liu2024gsbm}.
For path-space policy optimization, the GSB objective can be written as
\begin{equation}
    P^*
    =
    \arg\max_{P:\,P_0=\mu_{\mathrm{init}}}
    \left\{
        \mathbb E_{\tau\sim P}
        [U(a_1)]
        -
        \alpha
        D_{\mathrm{KL}}
        \left(
            P
            \|
            P_{\mathrm{ref}}
        \right)
    \right\},
    \label{eq:prelim_gsb_problem}
\end{equation}
where $P_{\mathrm{ref}}$ is a reference path measure, $P_0=\mu_{\mathrm{init}}$ fixes the simple prior endpoint, and $U(a_1)$ softly shapes the terminal action distribution instead of prescribing it as a hard marginal constraint.
For on-policy generative RL, the reference process is the old path policy and $U$ is the old-policy advantage of the executed action.

In implementation, the continuous process is discretized on a grid
\begin{equation}
    0=t_0<t_1<\cdots<t_N=1,
    \label{eq:prelim_time_grid}
\end{equation}
with $a^{(n)}\approx a_{t_n}$.
A simple Euler transition is
\begin{equation}
    a^{(n+1)}
    =
    a^{(n)}
    +
    (t_{n+1}-t_n)
    f_\theta(a^{(n)},t_n,s)
    +
    \sigma_{t_n}\sqrt{t_{n+1}-t_n}\,\epsilon_n,
    \qquad
    \epsilon_n\sim\mathcal N(0,I).
    \label{eq:prelim_discrete_transition}
\end{equation}
For the discrete path
\begin{equation}
    \tau:=a^{(0:N)}=(a^{(0)},a^{(1)},\ldots,a^{(N)}),
    \label{eq:prelim_path}
\end{equation}
the policy induces the path density
\begin{equation}
    P_\theta(a^{(0:N)}\mid s)
    =
    \mu_{\mathrm{init}}(a^{(0)})
    \prod_{n=0}^{N-1}
    p_\theta(a^{(n+1)}\mid a^{(n)},s).
    \label{eq:prelim_path_measure}
\end{equation}
This GSB view motivates GSB-MDPO in Section~\ref{sec:method}: mirror descent is lifted from terminal action distributions to generation path measures.

\section{Method}
\label{sec:method}

The central idea of \emph{GSB-MDPO} is to regard a generative policy not merely as a terminal action distribution, but as a conditional path measure over the full stochastic generation process.
This path-space view allows us to define proximal policy improvement directly over generation trajectories while reusing old-policy rollout paths through importance sampling.
Our formulation builds on mirror-descent policy optimization~\citep{geist2019regularized,tomar2020mdpo} and the generalized Schr\"odinger bridge view of stochastic generation as path-space optimization~\citep{leonard2013schrodinger,liu2024gsbm}.

\subsection{GSB-MDPO: Mirror Descent in Path Space}
\label{subsec:gsb_mdpo}

Let $P_\theta(\tau\mid s)$ denote the path measure induced by a generative policy, where $\tau=a^{(0:N)}$, $a^{(0)}\sim\mu_{\mathrm{init}}$ is the prior sample, and $a^{(N)}$ is the terminal action executed in the environment.
As introduced in Section~\ref{subsec:prelim_path_gsb}, the terminal action distribution is the marginal
\begin{equation}
    \pi_\theta(a^{(N)}\mid s)
    =
    \int
    P_\theta(a^{(0:N)}\mid s)
    \, da^{(0:N-1)}.
    \label{eq:method_terminal_marginal}
\end{equation}
Thus, for any advantage depending only on the executed action,
\begin{equation}
    \mathbb E_{a^{(N)}\sim \pi_\theta(\cdot\mid s)}
    [A(s,a^{(N)})]
    =
    \mathbb E_{\tau\sim P_\theta(\cdot\mid s)}
    [A(s,a^{(N)})].
    \label{eq:method_path_terminal_equivalence}
\end{equation}
Let $P_k:=P_{\theta_k}$ be the old path policy used to collect rollouts and let $A_k(s,a^{(N)})$ be the old-policy advantage estimate.
GSB-MDPO performs mirror descent over path measures:
\begin{equation}
    P_{k+1}
    =
    \arg\max_P
    \mathbb E_{s\sim d_k}
    \left[
        \mathbb E_{\tau\sim P(\cdot\mid s)}[A_k(s,a^{(N)})]
        -
        \alpha
        D_{\mathrm{KL}}\!\left(P(\cdot\mid s)\|P_k(\cdot\mid s)\right)
    \right],
    \label{eq:gsb_mdpo_objective}
\end{equation}
where $d_k$ is the old state distribution and $\alpha>0$ controls the mirror-descent step size.
The reward-improvement term acts on the terminal action, while the proximal regularizer acts on the full generation path.

\begin{proposition}[Path KL controls terminal KL]
\label{prop:path_kl_upper_bound}
Let $P$ and $Q$ be path measures over $\tau=a^{(0:N)}$, with terminal marginals $\pi_P$ and $\pi_Q$ over $a^{(N)}$.
Then
\begin{equation}
\begin{aligned}
    D_{\mathrm{KL}}(P\|Q)
    &\ge D_{\mathrm{KL}}(\pi_P\|\pi_Q)
\end{aligned}
\label{eq:path_kl_decomposition}
\end{equation}
\end{proposition}

The proof is given in Appendix~\ref{app:proof_path_kl}.
This is the KL chain rule, equivalently an instance of the data processing inequality~\citep{cover1999elements}.
Related auxiliary-variable bounds appear in variational inference and generative modeling~\citep{agakov2004auxiliary,tran2015copula,ranganath2016hierarchical,maaloe2016auxiliary,arenz2018efficient}, and have recently been used for diffusion-policy entropy bounds~\citep{celik2025dime}.
Here, the same structure shows that path-space regularization controls both the terminal marginal and the conditional generation dynamics.

\begin{proposition}[Expected old-advantage improvement]
\label{prop:advantage_improvement}
Let $P_{k+1}(\cdot\mid s)$ solve Eq.~\eqref{eq:gsb_mdpo_objective} for each state $s$. Then
\begin{equation}
    \mathbb E_{s\sim d_k,\,\tau\sim P_{k+1}(\cdot\mid s)}
    \left[
        A_k(s,a^{(N)})
    \right]
    \ge
    \mathbb E_{s\sim d_k,\,\tau\sim P_k(\cdot\mid s)}
    \left[
        A_k(s,a^{(N)})
    \right].
    \label{eq:advantage_improvement_state_average}
\end{equation}
\end{proposition}

The proof is given in Appendix~\ref{app:proof_advantage_improvement}.
This result states that the exact path-space mirror-descent update improves the expected old-policy advantage relative to the old path policy under the fixed old-state distribution.

\begin{proposition}[Optimal GSB-MDPO path measure]
\label{prop:optimal_gsb_mdpo}
For fixed $s$, the solution of Eq.~\eqref{eq:gsb_mdpo_objective} satisfies
\begin{equation}
    P^*(\tau\mid s)
    \propto
    P_k(\tau\mid s)
    \exp\!\left(\frac{A_k(s,a^{(N)})}{\alpha}\right).
    \label{eq:optimal_path_solution}
\end{equation}
Since $A_k$ depends only on the terminal action, this implies
\begin{align}
    \pi^*(a^{(N)}\mid s)
    &\propto
    \pi_k(a^{(N)}\mid s)
    \exp\!\left(\frac{A_k(s,a^{(N)})}{\alpha}\right),
    \label{eq:optimal_terminal_marginal}\\
    P^*(a^{(0:N-1)}\mid a^{(N)},s)
    &=
    P_k(a^{(0:N-1)}\mid a^{(N)},s).
    \label{eq:optimal_conditional_path}
\end{align}
A toy visualization of this exponential-tilting solution is provided in Appendix~\ref{app:toy_analysis}.
\end{proposition}

The proof is given in Appendix~\ref{app:proof_optimal_solution}.
Proposition~\ref{prop:optimal_gsb_mdpo} shows that the exact update increases the probability of high-advantage terminal actions while preserving the old conditional path structure given the terminal action.
This mirrors action-space KL-regularized policy improvement~\citep{geist2019regularized,tomar2020mdpo}, but applies it to path measures rather than one-step action densities.

\subsection{Importance-Sampled Path-KL Objective}
\label{subsec:is_path_kl_objective}

The exact objective in Eq.~\eqref{eq:gsb_mdpo_objective} involves expectations under $P_\theta$, while on-policy rollouts are collected from $P_k$.
For a stored path $\tau$, define the exact path ratio
\begin{equation}
    r_\theta(s,\tau)
    =
    \frac{P_\theta(\tau\mid s)}{P_k(\tau\mid s)}
    =
    \prod_{n=0}^{N-1}
    \frac{
        p_\theta(a^{(n+1)}\mid a^{(n)},s)
    }{
        p_k(a^{(n+1)}\mid a^{(n)},s)
    }.
    \label{eq:path_ratio}
\end{equation}
For any path-level function $F$,
\begin{equation}
    \mathbb E_{\tau\sim P_\theta(\cdot\mid s)}[F(s,\tau)]
    =
    \mathbb E_{\tau\sim P_k(\cdot\mid s)}[r_\theta(s,\tau)F(s,\tau)].
    \label{eq:is_identity_general}
\end{equation}
Taking $F(s,\tau)=A_k(s,a^{(N)})$ gives the importance-sampled advantage term.

The path KL also admits a drift-MSE form under the SDE view in Section~\ref{subsec:prelim_path_gsb}.
For two policies sharing the same initial prior $\mu_{\mathrm{init}}$ and diffusion coefficient $\sigma_t$, with drifts $f_\theta$ and $f_k$, Girsanov's theorem gives~\citep{oksendal2003stochastic}
\begin{equation}
    D_{\mathrm{KL}}(P_\theta\|P_k)
    =
    \mathbb E_{\tau\sim P_\theta}
    \left[
        \int_0^1
        \frac{\|f_\theta(a_t,t,s)-f_k(a_t,t,s)\|_2^2}{2\sigma_t^2}
        dt
    \right],
    \label{eq:path_kl_drift_energy}
\end{equation}
where conditioning on $s$ is omitted when clear.
After discretization, the path-KL drift cost is
\begin{equation}
    C_{\mathrm{path}}(\theta,\theta_k;s,\tau)
    =
    \sum_{n=0}^{N-1}
    \frac{\Delta t_n}{2\sigma(t_n)^2}
    \|f_\theta(a^{(n)},t_n,s)-f_k(a^{(n)},t_n,s)\|_2^2,
    \qquad
    \Delta t_n=t_{n+1}-t_n.
    \label{eq:path_cost}
\end{equation}
Thus the MSE-style drift discrepancy is not an ad-hoc penalty; it is the discretized path-KL form for stochastic generators with shared diffusion coefficient.
Using Eq.~\eqref{eq:is_identity_general}, the practical objective folds this cost into an effective advantage:
\begin{equation}
    \mathcal L_{\mathrm{MDPO}}(\theta)
    =
    -
    \mathbb E_{s\sim d_k,\tau\sim P_k}
    \left[
        r_\theta(s,\tau)
        \big(A_k(s,a^{(N)})-\lambda C_{\mathrm{path}}(\theta,\theta_k;s,\tau)\big)
    \right].
    \label{eq:mdpo_is_loss}
\end{equation}
This matches the on-policy mirror-descent objective because both the advantage and path-KL cost are evaluated under the current path measure and then rewritten on old-policy paths by importance sampling.
However, the exact path ratio is a product of stepwise likelihood ratios and can have high variance when the current generator moves far from the old one.
For this reason, our implementation uses the stabilized path-ratio estimator described in Section~\ref{subsec:ratio_clipping}.

\subsection{Entropy-Preserving Reference Regularization}
\label{subsec:entropy_regularization}

In some environments, it is useful to regularize the path policy not only toward the old policy but also toward a simple reference process.
Let $P_{\mathrm{ref}}(\cdot\mid s)$ denote a reference path measure, such as a zero-drift noising process.
We consider
{\small
\begin{equation}
\begin{aligned}
    \max_P\;
    \mathbb E_{s\sim d_k}
    \Big[
        \mathbb E_{\tau\sim P(\cdot\mid s)}[A_k(s,a^{(N)})]
        -
        \alpha D_{\mathrm{KL}}(P(\cdot\mid s)\|P_k(\cdot\mid s))
        -
        \beta D_{\mathrm{KL}}(P(\cdot\mid s)\|P_{\mathrm{ref}}(\cdot\mid s))
    \Big].
\end{aligned}
    \label{eq:entropy_extension_objective}
\end{equation}
}
The two KL terms combine into a single KL toward
\begin{equation}
    P_\eta(\tau\mid s)
    =
    \frac{1}{Z_\eta(s)}
    P_k(\tau\mid s)^{1-\eta}
    P_{\mathrm{ref}}(\tau\mid s)^\eta,
    \qquad
    \eta=\frac{\beta}{\alpha+\beta},
    \label{eq:composite_reference}
\end{equation}
since
\begin{equation}
    \alpha D_{\mathrm{KL}}(P\|P_k)+\beta D_{\mathrm{KL}}(P\|P_{\mathrm{ref}})
    =
    (\alpha+\beta)D_{\mathrm{KL}}(P\|P_\eta)
    -
    (\alpha+\beta)\log Z_\eta(s).
    \label{eq:composite_kl_identity}
\end{equation}
Reference-process regularization is common in Schr\"odinger bridge formulations~\citep{leonard2013schrodinger,liu2024gsbm} and has recently appeared in maximum-entropy generative RL~\citep{celik2025dime,lv2026flac}.

In drift space, completing the square yields the mixed anchor
\begin{equation}
    f_\eta=(1-\eta)f_k+\eta f_{\mathrm{ref}}.
    \label{eq:mixed_anchor}
\end{equation}
In our implementation, $f_{\mathrm{ref}}$ is the zero drift, so the per-path cost becomes
\begin{equation}
    C_{\eta}(\theta,\theta_k;s,\tau)
    =
    \sum_{n=0}^{N-1}
    \frac{\Delta t_n}{2\sigma(t_n)^2}
    \|f_\theta(a^{(n)},t_n,s)-f_\eta(a^{(n)},t_n,s)\|_2^2,
    \qquad
    \Delta t_n=t_{n+1}-t_n.
    \label{eq:mixed_anchor_regularizer}
\end{equation}
When reference regularization is disabled, $\eta=0$ and $C_\eta=C_{\mathrm{path}}$.

\subsection{Step- and Path-Level Ratio Clipping}
\label{subsec:ratio_clipping}

The theoretical objective uses the exact path ratio $r_\theta(s,\tau)$.
In practice, the path ratio is a product of generation-step likelihood ratios, so numerical errors can accumulate across the generation path.
For each generation step, let
\[
    \Delta\ell_n
    =
    \log p_\theta(a^{(n+1)}\mid a^{(n)},s)
    -
    \log p_k(a^{(n+1)}\mid a^{(n)},s),
    \qquad n=0,\ldots,N-1.
\]
We use the clipped path ratio
\begin{equation}
    \widetilde r_\theta(s,\tau)
    =
    \exp\!\left[
    \mathrm{clip}\!\left(
    \sum_{n=0}^{N-1}
    \mathrm{clip}(\Delta\ell_n,-c_{\mathrm{step}},c_{\mathrm{step}}),
    -c_{\mathrm{path}},c_{\mathrm{path}}
    \right)
    \right].
    \label{eq:clipped_path_ratio}
\end{equation}
The implemented policy loss is
\begin{equation}
    \widehat{\mathcal L}_{\mathrm{MDPO}}(\theta)
    =
    -
    \mathbb E_{s\sim d_k,\tau\sim P_k}
    \left[
        \widetilde r_\theta(s,\tau)
        \big(A_k(s,a^{(N)})-\lambda C_\eta(\theta,\theta_k;s,\tau)\big)
    \right].
    \label{eq:clipped_mdpo_loss}
\end{equation}
The step-level and path-level clipping operations are applied to the accumulated path log-ratio before forming the ratio used in the policy objective.
This is a numerical stabilization mechanism for the accumulated path likelihood ratio, not a PPO-style clipped surrogate: PPO clips a terminal action ratio, whereas GSB-MDPO clips stepwise generation log-ratios and their accumulated path log-ratio to control path-ratio variance during minibatch optimization.
When the likelihood ratio is clipped, gradients through the ratio are suppressed, but the drift cost $C_\eta(\theta,\theta_k;s,\tau)$ still contributes gradients through the effective advantage term in Eq.~\eqref{eq:clipped_mdpo_loss}.
Thus, clipping stabilizes the importance weight without removing the path-KL drift regularization signal.

Algorithm~\ref{alg:gsb_mdpo} summarizes the implementation.
All policy updates reuse paths sampled from the old policy; during optimization, we only recompute current path likelihoods and drift values on the stored rollout paths.

\begin{algorithm}[t]
\caption{GSB-MDPO}
\label{alg:gsb_mdpo}
\begin{algorithmic}[1]
\Require Actor $\theta$, critic $\phi$, coefficients $\lambda,\eta$, clipping thresholds $c_{\mathrm{step}},c_{\mathrm{path}}$
\For{each on-policy iteration $k=0,1,2,\ldots$}
    \State Set old actor $\theta_k\leftarrow\theta$
    \State Collect rollouts with $P_k=P_{\theta_k}$ and store paths, old log-likelihoods, and old drifts
    \State Compute advantages and value targets
    \For{each minibatch}
        \State Recompute current log-likelihoods and drifts on stored paths
        \State Compute path cost $C_\eta$ using Eq.~\eqref{eq:mixed_anchor_regularizer}
        \State Compute clipped path ratio $\widetilde r_\theta$ using Eq.~\eqref{eq:clipped_path_ratio}
        \State Update actor with Eq.~\eqref{eq:clipped_mdpo_loss}
        \State Update critic $\phi$ by minimizing
        $\mathcal L_V(\phi)=
        \mathbb E_{s\sim\mathcal D_k}
        \left[
        \left(
        V_\phi(s)-\hat V_k(s)
        \right)^2
        \right]$
    \EndFor
\EndFor
\end{algorithmic}
\end{algorithm}

\section{Experiments}
\label{sec:experiments}

We evaluate GSB-MDPO on two continuous-control benchmark suites:
(i) eight Playground tasks from the FPO benchmark~\citep{mcallister2025fpo},
built on MuJoCo Playground~\citep{zakka2025mujoco}; and
(ii) six Gym-MuJoCo tasks implemented in our on-policy FlowRL-based
pipeline~\citep{gao2026flowrl}, using the Gymnasium interface
and the MuJoCo physics engine~\citep{towers2024gymnasium,todorov2012mujoco}.
We compare five methods: Gaussian PPO~\citep{schulman2017ppo}, DPPO~\citep{ren2024dppo}, FPO~\citep{mcallister2025fpo}, GenPO~\citep{ding2025genpo}, and GSB-MDPO.
Within each benchmark, each method uses one fixed hyperparameter configuration across all tasks, without task-specific tuning, except for a memory-related GenPO batch-size adjustment on Humanoid-v5 described in Appendix~\ref{app:implementation}.
For Gym-MuJoCo, all methods are trained for $40$M environment interactions on the same six tasks.
Main benchmark curves report the mean over five seeds, with shaded regions showing one standard deviation.
Ablation curves report the mean over two seeds, with shaded regions showing the min--max envelope.
Implementation details and hyperparameters are provided in Appendix~\ref{app:implementation}.

\subsection{Main Results on Playground}
\label{subsec:playground_results}

\begin{figure}[t]
    \centering
    \includegraphics[width=\linewidth]{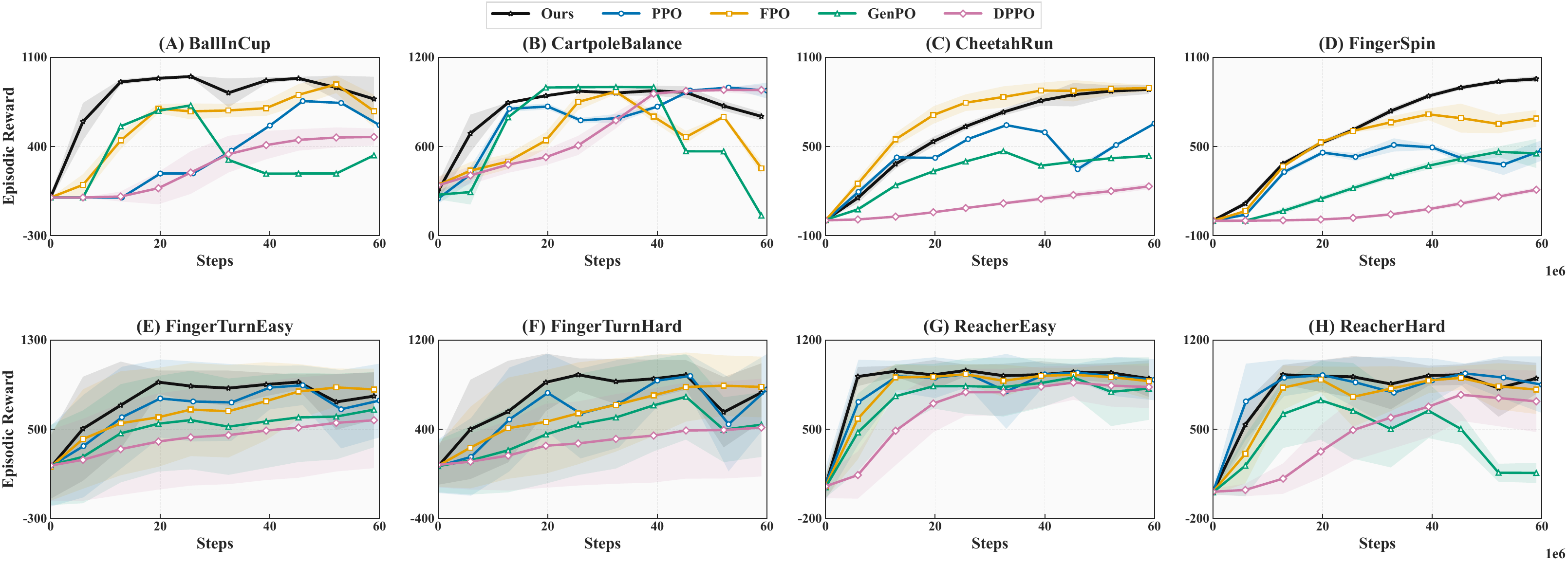}
    \caption{
    Playground results. Curves show mean $\pm$ one standard deviation over five seeds.
    }
    \label{fig:playground_main}
    \vspace{-0.5cm}
\end{figure}

Figure~\ref{fig:playground_main} reports results on the Playground suite.
GSB-MDPO obtains competitive or higher returns across the eight tasks, with clear gains on BallInCup, FingerSpin, FingerTurnEasy, and FingerTurnHard.
On CheetahRun and the two reaching tasks, multiple methods reach comparable final returns, and GSB-MDPO remains among the leading methods.

These results suggest that the gains are not merely due to using a multi-step generative policy.
Compared with DPPO, FPO, and GenPO, GSB-MDPO regularizes the full generation path, which provides a more direct proximal update for on-policy training.

\subsection{Main Results on Gym-MuJoCo}
\label{subsec:gym_results}

\begin{figure}[t]
    \centering
    \includegraphics[width=\linewidth]{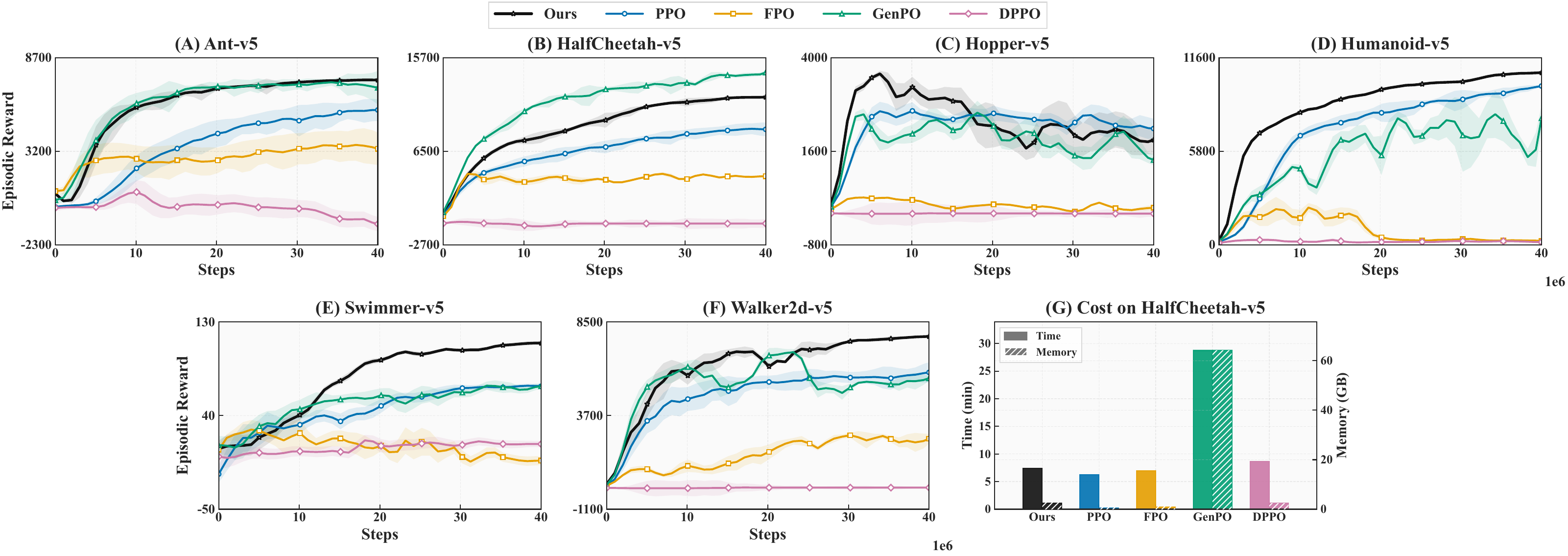}
    \caption{Gym-MuJoCo results. Panels (A)--(F) show learning curves on six tasks, with mean $\pm$ one standard deviation over five seeds. Panel (G) reports representative training time and GPU memory on HalfCheetah-v5.}
    \label{fig:gym_main}
\end{figure}

Figure~\ref{fig:gym_main} reports results on Gym-MuJoCo.
GSB-MDPO performs strongly across the six locomotion tasks and is especially effective on higher-dimensional environments such as Ant-v5 and Humanoid-v5, where both state and action spaces are larger.
It also obtains strong final performance on Swimmer-v5 and Walker2d-v5, and remains competitive on Hopper-v5.
On HalfCheetah-v5, GenPO achieves the highest return, while GSB-MDPO remains competitive under the same benchmark-level hyperparameter configuration used for all Gym-MuJoCo tasks, reflecting the tradeoff of using one fixed configuration across heterogeneous environments.

Panel (G) shows that GSB-MDPO does not introduce prohibitive training overhead.
On HalfCheetah-v5, its training time is close to PPO/FPO/DPPO and substantially lower than GenPO, while its memory usage remains comparable to other path-based generative baselines.
Thus, the Gym-MuJoCo results support path-space mirror descent as an effective and practical optimization principle for on-policy generative policies across a separate FlowRL-based training stack.

\subsection{Further Analysis}
\label{subsec:ablations}

\begin{figure}[H]
    \centering
    \includegraphics[width=\linewidth]{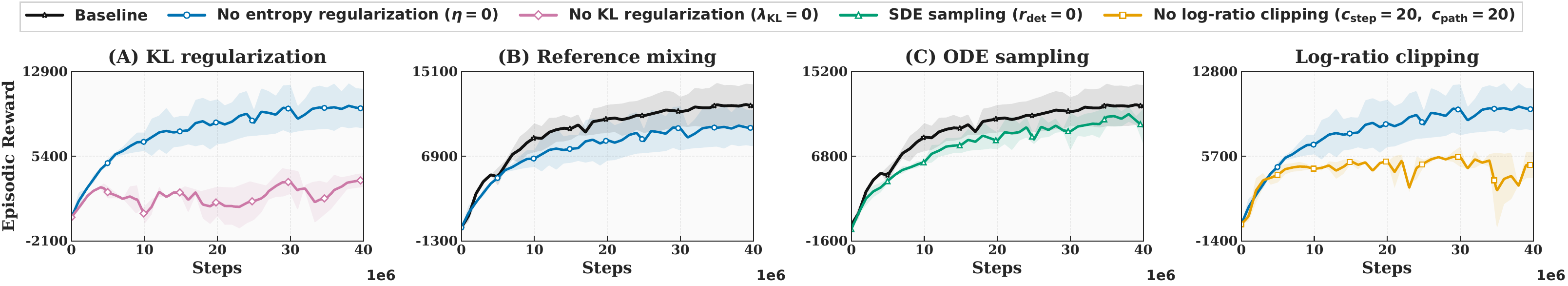}
    \caption{
    Ablations on HalfCheetah-v5. Curves show the mean and min--max envelope over two seeds.
    }
    \label{fig:ablation}
\end{figure}

Figure~\ref{fig:ablation} evaluates four components of GSB-MDPO on HalfCheetah-v5.
All variants are trained under the same benchmark-level configuration, with only the tested component changed.
For the KL and Log-ratio clipping ablation, we compare against the $\eta=0$ version of GSB-MDPO so that the effect of the path-KL coefficient is isolated from reference mixing.

\textbf{Path-KL regularization (Fig.~\ref{fig:ablation}A).}
The path-KL coefficient provides the proximal constraint required by the path-space mirror-descent update.
Setting this coefficient to zero substantially degrades final performance, indicating that the drift-space penalty is an essential part of the optimization objective rather than an auxiliary stabilizer.

\textbf{Reference mixing (Fig.~\ref{fig:ablation}B).}
A positive reference-mixing coefficient improves performance over the $\eta=0$ variant.
This indicates that reference mixing acts as an entropy-like regularizer, helping maintain stochasticity in the generative path distribution and preventing premature policy concentration.

\textbf{Deterministic ODE evaluation (Fig.~\ref{fig:ablation}C).}
This result supports using the deterministic path at evaluation time: the policy is trained with stochastic generation paths, while evaluation removes sampling noise by following the learned deterministic drift.

\textbf{Log-ratio clipping (Fig.~\ref{fig:ablation}D).}
Step- and path-level log-ratio clipping improves the stability of importance-sampled path-ratio optimization.
Without clipping, accumulated generation-step likelihood ratios produce a less reliable update.

% Overall, the ablations support the main design choices of GSB-MDPO: path-level proximal regularization, reference-process mixing, deterministic evaluation, and stabilized path ratios each contribute to robust empirical performance.

\begin{wrapfigure}{r}{0.35\textwidth}
    \centering
    \vspace{-12pt}
    \includegraphics[width=1.0\linewidth]{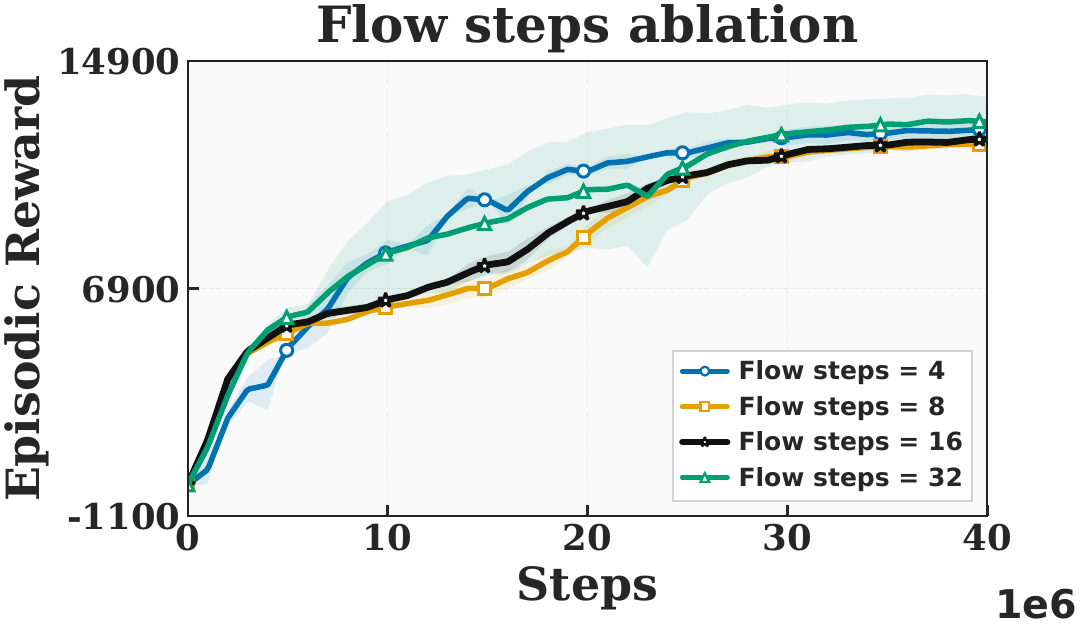}
    \caption{Ablation of the number of generation steps for GSB-MDPO on HalfCheetah-v5. Solid curves show the mean over two seeds, and shaded regions show the min--max envelope across seeds.}
    \vspace{-12pt}
    \label{fig:flow_steps_ablation}
\end{wrapfigure}

\textbf{Sensitivity to Generation Steps.} We additionally study the sensitivity of GSB-MDPO to the number of generation steps.
Figure~\ref{fig:flow_steps_ablation} reports results on HalfCheetah-v5 with $4$, $8$, $16$, and $32$ flow steps.
All variants are trained for $40$M environment interactions and averaged over two random seeds.

The results show that GSB-MDPO is not highly sensitive to the exact number of generation steps on this task.
All four settings reach similar final performance by the end of training.
Using more steps can slightly smooth the later-stage improvement, while fewer steps such as $4$ or $8$ remain competitive and sometimes learn faster early in training.
This suggests that the path-space MDPO objective is compatible with both short and moderately long generative paths, and that reducing the number of generation steps may be a practical way to trade off sampling cost and performance.

\section{Conclusion and Limitations}
\label{sec:conclusion_limitations}

GSB-MDPO connects generalized Schr\"odinger bridge modeling with mirror descent policy optimization to derive a path-space proximal update for multi-step generative policies. 
Instead of relying on generally intractable terminal action densities, it regularizes the conditional generation path measure. The resulting path KL controls the terminal action KL, can be estimated from on-policy data, and reduces to a drift-MSE penalty under shared diffusion coefficients.
Despite optimizing multi-step generation paths, GSB-MDPO achieves PPO-comparable efficiency by avoiding additional current-policy sampling.
Experiments on Playground and Gym-MuJoCo, together with ablations, support the effectiveness of path-space regularization for on-policy generative reinforcement learning.

Despite its strong performance, GSB-MDPO currently uses a fixed scalar noise schedule shared across action dimensions.
Although effective on normalized continuous-control benchmarks, this design may be suboptimal for heterogeneous actuators that require state-dependent stochasticity, a limitation also noted in related bridge-based maximum-entropy methods~\citep{lv2026flac}.
Future work should explore adaptive noise schedules and fewer-step path-space updates for broader control domains.

%%%%%%%%%%%%%%%%%%%%%%%%%%%%%%%%%%%%%%%%%%%%%%%%%%%%%%%%%%%%
% Acknowledgments
%%%%%%%%%%%%%%%%%%%%%%%%%%%%%%%%%%%%%%%%%%%%%%%%%%%%%%%%%%%%

% Do not include acknowledgments in the anonymized submission.
% The ack environment is hidden automatically in anonymous mode.
%
% \begin{ack}
% We thank ...
% \end{ack}

%%%%%%%%%%%%%%%%%%%%%%%%%%%%%%%%%%%%%%%%%%%%%%%%%%%%%%%%%%%%
% References
%%%%%%%%%%%%%%%%%%%%%%%%%%%%%%%%%%%%%%%%%%%%%%%%%%%%%%%%%%%%
\clearpage
\bibliographystyle{plainnat}
\bibliography{refs}

\clearpage
%%%%%%%%%%%%%%%%%%%%%%%%%%%%%%%%%%%%%%%%%%%%%%%%%%%%%%%%%%%%
% Appendix
%%%%%%%%%%%%%%%%%%%%%%%%%%%%%%%%%%%%%%%%%%%%%%%%%%%%%%%%%%%%

\appendix

\section{Proofs}
\label{app:proofs}

\subsection{Proof of the Path-KL Decomposition}
\label{app:proof_path_kl}

Let $P$ and $Q$ be path measures over $\tau=a^{(0:N)}$.
Factorizing both measures by the terminal action gives
\begin{align}
    P(a^{(0:N)})
    &=
    \pi_P(a^{(N)})
    P(a^{(0:N-1)}\mid a^{(N)}),
    \\
    Q(a^{(0:N)})
    &=
    \pi_Q(a^{(N)})
    Q(a^{(0:N-1)}\mid a^{(N)}).
\end{align}
Then
\begin{align}
    D_{\mathrm{KL}}(P\|Q)
    &=
    \mathbb E_{\tau\sim P}
    \left[
        \log\frac{P(a^{(0:N)})}{Q(a^{(0:N)})}
    \right]
    \\
    &=
    \mathbb E_{\tau\sim P}
    \left[
        \log\frac{\pi_P(a^{(N)})}{\pi_Q(a^{(N)})}
        +
        \log
        \frac{
            P(a^{(0:N-1)}\mid a^{(N)})
        }{
            Q(a^{(0:N-1)}\mid a^{(N)})
        }
    \right]
    \\
    &=
    D_{\mathrm{KL}}(\pi_P\|\pi_Q)
    +
    \mathbb E_{a^{(N)}\sim \pi_P}
    \left[
        D_{\mathrm{KL}}
        \left(
            P(a^{(0:N-1)}\mid a^{(N)})
            \|
            Q(a^{(0:N-1)}\mid a^{(N)})
        \right)
    \right]
    \\
    &\ge
    D_{\mathrm{KL}}(\pi_P\|\pi_Q).
\end{align}

\subsection{Proof of Expected Old-Advantage Improvement}
\label{app:proof_advantage_improvement}

For fixed $s$, define
\begin{equation}
    \mathcal J_s(P)
    =
    \mathbb E_{\tau\sim P(\cdot\mid s)}
    [A_k(s,a^{(N)})]
    -
    \alpha
    D_{\mathrm{KL}}
    \left(
        P(\cdot\mid s)
        \|
        P_k(\cdot\mid s)
    \right).
    \label{eq:proof_js_def}
\end{equation}
Since $P_{k+1}$ maximizes $\mathcal J_s$ and $P_k$ is feasible,
\begin{align}
    \mathbb E_{\tau\sim P_{k+1}}
    [A_k(s,a^{(N)})]
    -
    \alpha
    D_{\mathrm{KL}}(P_{k+1}\|P_k)
    &=
    \mathcal J_s(P_{k+1})
    \label{eq:proof_adv_start}\\
    &\ge
    \mathcal J_s(P_k)
    \label{eq:proof_adv_optimality}\\
    &=
    \mathbb E_{\tau\sim P_k}
    [A_k(s,a^{(N)})].
    \label{eq:proof_adv_old}
\end{align}
Therefore,
\begin{align}
    \mathbb E_{\tau\sim P_{k+1}}
    [A_k(s,a^{(N)})]
    &\ge
    \mathbb E_{\tau\sim P_k}
    [A_k(s,a^{(N)})]
    +
    \alpha
    D_{\mathrm{KL}}(P_{k+1}\|P_k)
    \label{eq:proof_adv_stronger}\\
    &\ge
    \mathbb E_{\tau\sim P_k}
    [A_k(s,a^{(N)})].
    \label{eq:proof_adv_final}
\end{align}
Taking expectation over $s\sim d_k$ gives the state-averaged result.

\subsection{Proof of the Optimal GSB-MDPO Path Measure}
\label{app:proof_optimal_solution}

Fix $s$ and write $\tau=a^{(0:N)}$. The per-state optimization problem is
\begin{align}
    \max_{P}
    \quad
    &
    \int
    P(\tau)
    A_k(s,a^{(N)})\,d\tau
    -
    \alpha
    \int
    P(\tau)
    \log
    \frac{P(\tau)}{P_k(\tau)}
    d\tau,
    \label{eq:proof_opt_problem}\\
    \mathrm{s.t.}
    \quad
    &
    \int P(\tau)d\tau=1.
    \label{eq:proof_opt_constraint}
\end{align}
Introducing a Lagrange multiplier $\lambda$ gives
\begin{align}
    \mathcal L(P,\lambda)
    &=
    \int
    P(\tau)
    A_k(s,a^{(N)})\,d\tau
    -
    \alpha
    \int
    P(\tau)
    \log
    \frac{P(\tau)}{P_k(\tau)}
    d\tau
    \label{eq:proof_lagrangian_first}\\
    &\quad+
    \lambda
    \left(
        \int P(\tau)d\tau-1
    \right).
    \label{eq:proof_lagrangian_second}
\end{align}
The first-order condition and its rearranged form are
\begin{align}
    \frac{\delta \mathcal L}{\delta P(\tau)}
    &=
    A_k(s,a^{(N)})
    -
    \alpha
    \left(
        \log
        \frac{P^*(\tau)}{P_k(\tau)}
        +
        1
    \right)
    +
    \lambda
    =
    0,
    \label{eq:proof_stationarity}\\
    \log
    \frac{P^*(\tau)}{P_k(\tau)}
    &=
    \frac{A_k(s,a^{(N)})}{\alpha}
    +
    \frac{\lambda}{\alpha}
    -
    1.
    \label{eq:proof_log_ratio_solution}
\end{align}
Absorbing the constant $\exp(\lambda/\alpha-1)$ into the normalizer yields
\begin{align}
    P^*(\tau\mid s)
    &=
    \frac{1}{Z(s)}
    P_k(\tau\mid s)
    \exp
    \left(
        \frac{A_k(s,a^{(N)})}{\alpha}
    \right),
    \label{eq:proof_optimal_path}\\
    Z(s)
    &=
    \int
    P_k(\tau\mid s)
    \exp
    \left(
        \frac{A_k(s,a^{(N)})}{\alpha}
    \right)
    d\tau.
    \label{eq:proof_partition_integral}
\end{align}
Equivalently,
\begin{equation}
    Z(s)
    =
    \mathbb E_{\tau\sim P_k(\cdot\mid s)}
    \left[
        \exp
        \left(
            \frac{A_k(s,a^{(N)})}{\alpha}
        \right)
    \right].
    \label{eq:proof_partition_expectation}
\end{equation}
Since the tilt depends only on the terminal action $a^{(N)}$, the joint optimizer factorizes as
\begin{align}
    P^*(a^{(0:N)}\mid s)
    &=
    \frac{1}{Z(s)}
    \pi_k(a^{(N)}\mid s)
    P_k(a^{(0:N-1)}\mid a^{(N)},s)
    \exp
    \left(
        \frac{A_k(s,a^{(N)})}{\alpha}
    \right),
    \label{eq:proof_optimal_factorized}\\
    \pi^*(a^{(N)}\mid s)
    &=
    \int
    P^*(a^{(0:N)}\mid s)
    da^{(0:N-1)}
    \label{eq:proof_optimal_marginal_integral}\\
    &=
    \frac{1}{Z(s)}
    \pi_k(a^{(N)}\mid s)
    \exp
    \left(
        \frac{A_k(s,a^{(N)})}{\alpha}
    \right),
    \label{eq:proof_optimal_marginal}\\
    P^*(a^{(0:N-1)}\mid a^{(N)},s)
    &=
    \frac{
        P^*(a^{(0:N)}\mid s)
    }{
        \pi^*(a^{(N)}\mid s)
    }
    \label{eq:proof_optimal_conditional_ratio}\\
    &=
    P_k(a^{(0:N-1)}\mid a^{(N)},s).
    \label{eq:proof_optimal_conditional}
\end{align}

\subsection{Importance-Sampling Identities}
\label{app:proof_importance_sampling}

For any path-level function $F(s,\tau)$,
\begin{align}
    \mathbb E_{\tau\sim P_\theta(\cdot\mid s)}
    \left[
        F(s,\tau)
    \right]
    &=
    \int
    P_\theta(\tau\mid s)
    F(s,\tau)
    d\tau
    \label{eq:proof_is_start}\\
    &=
    \int
    P_k(\tau\mid s)
    \frac{P_\theta(\tau\mid s)}{P_k(\tau\mid s)}
    F(s,\tau)
    d\tau
    \label{eq:proof_is_change_measure}\\
    &=
    \mathbb E_{\tau\sim P_k(\cdot\mid s)}
    \left[
        r_\theta(s,\tau)F(s,\tau)
    \right],
    \label{eq:proof_is_identity}
\end{align}
which recovers Eq.~\eqref{eq:is_identity_general}.
Taking $F(s,\tau)=A_k(s,a^{(N)})$ gives the importance-sampled advantage term.
Similarly,
\begin{align}
    D_{\mathrm{KL}}
    \left(
        P_\theta(\cdot\mid s)
        \|
        P_k(\cdot\mid s)
    \right)
    &=
    \int
    P_\theta(\tau\mid s)
    \log
    \frac{P_\theta(\tau\mid s)}{P_k(\tau\mid s)}
    d\tau
    \label{eq:proof_is_kl_start}\\
    &=
    \int
    P_k(\tau\mid s)
    \frac{P_\theta(\tau\mid s)}{P_k(\tau\mid s)}
    \log
    \frac{P_\theta(\tau\mid s)}{P_k(\tau\mid s)}
    d\tau
    \label{eq:proof_is_kl_change_measure}\\
    &=
    \mathbb E_{\tau\sim P_k(\cdot\mid s)}
    \left[
        r_\theta(s,\tau)\log r_\theta(s,\tau)
    \right].
    \label{eq:proof_is_kl_identity}
\end{align}

\subsection{Path KL as Drift MSE}
\label{app:proof_path_kl_drift_mse}

Consider two path measures $P_\theta$ and $P_k$ induced by SDEs with the same initial prior $\mu_{\mathrm{init}}$ and the same diffusion coefficient:
\begin{align}
    da_t
    &=
    f_\theta(a_t,t,s)\,dt+\sigma_t\,dW_t,
    \qquad a_0\sim\mu_{\mathrm{init}},
    \label{eq:proof_sde_theta}\\
    da_t
    &=
    f_k(a_t,t,s)\,dt+\sigma_t\,dW_t,
    \qquad a_0\sim\mu_{\mathrm{init}}.
    \label{eq:proof_sde_old}
\end{align}
Under the standard absolute-continuity assumptions of Girsanov's theorem,
\begin{equation}
    D_{\mathrm{KL}}
    \left(
        P_\theta(\cdot\mid s)
        \|
        P_k(\cdot\mid s)
    \right)
    =
    \mathbb E_{\tau\sim P_\theta(\cdot\mid s)}
    \left[
        \int_0^1
        \frac{
            \|f_\theta(a_t,t,s)-f_k(a_t,t,s)\|_2^2
        }{
            2\sigma_t^2
        }
        dt
    \right].
    \label{eq:proof_path_kl_drift_energy}
\end{equation}
No initial KL term appears because the two path measures share the same initial prior.
A first-order discretization over the grid $0=t_0<\cdots<t_N=1$ gives
\begin{align}
    \int_0^1
    \frac{
        \|f_\theta(a_t,t,s)-f_k(a_t,t,s)\|_2^2
    }{
        2\sigma_t^2
    }
    dt
    &\approx
    \sum_{n=0}^{N-1}
    \frac{\Delta t_n}{2\sigma(t_n)^2}
    \left\|
        f_\theta(a^{(n)},t_n,s)
        -
        f_k(a^{(n)},t_n,s)
    \right\|_2^2
    \label{eq:proof_path_kl_discrete}\\
    &=
    C_{\mathrm{path}}(\theta,\theta_k;s,\tau),
    \label{eq:proof_path_cost}
\end{align}
where $\Delta t_n=t_{n+1}-t_n>0$.
This yields the drift-MSE path regularizer in Eq.~\eqref{eq:path_cost}.

\subsection{Composite KL and Mixed-Anchor Drift Penalty}
\label{app:proof_composite_kl}

For fixed $s$, write all path measures as conditional measures given $s$.
Let
\begin{align}
    \eta
    &=
    \frac{\beta}{\alpha+\beta},
    \label{eq:proof_eta_def}\\
    P_\eta(\tau\mid s)
    &=
    \frac{1}{Z_\eta(s)}
    P_k(\tau\mid s)^{1-\eta}
    P_{\mathrm{ref}}(\tau\mid s)^\eta.
    \label{eq:proof_composite_ref_def}
\end{align}
Then
\begin{equation}
    \log P_\eta(\tau\mid s)
    =
    (1-\eta)\log P_k(\tau\mid s)
    +
    \eta\log P_{\mathrm{ref}}(\tau\mid s)
    -
    \log Z_\eta(s).
    \label{eq:proof_log_composite_ref}
\end{equation}
Therefore,
\begin{align}
    &\alpha
    D_{\mathrm{KL}}
    \left(
        P(\cdot\mid s)
        \|
        P_k(\cdot\mid s)
    \right)
    +
    \beta
    D_{\mathrm{KL}}
    \left(
        P(\cdot\mid s)
        \|
        P_{\mathrm{ref}}(\cdot\mid s)
    \right)
    \label{eq:proof_composite_start}\\
    &=
    (\alpha+\beta)
    \mathbb E_{\tau\sim P(\cdot\mid s)}
    \left[
        \log P(\tau\mid s)
        -
        (1-\eta)\log P_k(\tau\mid s)
        -
        \eta\log P_{\mathrm{ref}}(\tau\mid s)
    \right]
    \label{eq:proof_composite_expand}\\
    &=
    (\alpha+\beta)
    \mathbb E_{\tau\sim P(\cdot\mid s)}
    \left[
        \log
        \frac{P(\tau\mid s)}{P_\eta(\tau\mid s)}
    \right]
    -
    (\alpha+\beta)\log Z_\eta(s)
    \label{eq:proof_composite_substitute}\\
    &=
    (\alpha+\beta)
    D_{\mathrm{KL}}
    \left(
        P(\cdot\mid s)
        \|
        P_\eta(\cdot\mid s)
    \right)
    -
    (\alpha+\beta)\log Z_\eta(s).
    \label{eq:proof_composite_kl}
\end{align}
For fixed $s$, $Z_\eta(s)$ is independent of the optimized conditional path measure $P(\cdot\mid s)$.
Since the mirror-descent update takes the outer expectation over the fixed old-policy state distribution $d_k$, the normalizer contributes a policy-independent constant to the fixed-$d_k$ objective.

For fixed path samples and shared diffusion coefficient, the pointwise weighted drift discrepancy satisfies
\begin{align}
    &\alpha\|f_\theta-f_k\|_2^2
    +
    \beta\|f_\theta-f_{\mathrm{ref}}\|_2^2
    \label{eq:proof_mixed_anchor_start}\\
    &=
    (\alpha+\beta)
    \left\|
        f_\theta
        -
        \frac{\alpha f_k+\beta f_{\mathrm{ref}}}{\alpha+\beta}
    \right\|_2^2
    +
    \frac{\alpha\beta}{\alpha+\beta}
    \|f_k-f_{\mathrm{ref}}\|_2^2.
    \label{eq:proof_mixed_anchor_square}
\end{align}
With
\begin{equation}
    f_\eta=(1-\eta)f_k+\eta f_{\mathrm{ref}},
    \label{eq:proof_mixed_anchor_def}
\end{equation}
the first term gives the mixed-anchor drift penalty.
When the drift loss is evaluated on fixed rollout paths, the residual term in Eq.~\eqref{eq:proof_mixed_anchor_square} does not depend on the optimized drift $f_\theta$ and can be omitted from the practical surrogate.
Discretizing the resulting drift discrepancy over $0=t_0<\cdots<t_N=1$ gives Eq.~\eqref{eq:mixed_anchor_regularizer}.

\section{Implementation Details}
\label{app:implementation}

\subsection{Hardware Configurations}
\label{app:hardware}

All experiments were carried out on a server equipped with two AMD EPYC 7763 64-Core processors.
Each socket contains 64 physical cores with 2 hardware threads per core, yielding up to 256 logical CPUs.
The maximum CPU frequency is approximately 3.53 GHz.
The system has a two-node NUMA topology, with CPUs distributed across the two sockets.

For GPU acceleration, we used 8 NVIDIA A800 80GB PCIe GPUs, each with 80 GB of memory.
The GPUs were operated with NVIDIA driver version 570.211.01 and CUDA 12.8.
Multi-Instance GPU (MIG) was disabled during the experiments.
The machine is equipped with 2.0 TiB of system memory and runs Ubuntu 24.04.3 LTS with Linux kernel 6.8.0.

\subsection{Codebases and Experimental Frameworks}
\label{app:codebases}

Our experiments are implemented on top of two public codebases.
For the Playground experiments, we use the official FPO repository, whose \texttt{playground/} implementation provides a JAX-based training pipeline for MuJoCo Playground and Brax~\citep{mcallister2025fpo}.
The FPO codebase already supports PPO, FPO, and DPPO within the same Playground training interface; DPPO is implemented as one mode of the FPO training script.
We reproduce these three baselines using the original Playground implementation and hyperparameter structure.
We further implement GenPO in the same FPO Playground pipeline by adapting the corresponding GenPO implementation logic from FlowRL.
Our GSB-MDPO agent is added to this same pipeline, so all Playground comparisons share the same environment interface, rollout collection procedure, training loop, and evaluation protocol.

For the Gym-MuJoCo experiments, we build on FlowRL, a modular JAX framework for reinforcement learning with diffusion and flow policies~\citep{gao2026flowrl}.
FlowRL provides implementations of PPO, DPPO, FPO, and GenPO, together with standardized infrastructure for diffusion/flow-policy RL.
The original FlowRL codebase includes on-policy IsaacLab launch scripts and off-policy Gym-MuJoCo launch scripts.
Since our Gym-MuJoCo comparison is on-policy, we implement an additional on-policy Gym-MuJoCo launcher following the FlowRL training interface.
We also implement GSB-MDPO in this FlowRL-based Gym-MuJoCo pipeline.
Due to interface differences between FPO Playground and FlowRL/Gym-MuJoCo, the Playground and Gym-MuJoCo implementations of GSB-MDPO are maintained separately, but they share the same core algorithmic logic: old-policy path rollout storage, current path likelihood recomputation, importance-sampled MDPO loss, and path-KL drift regularization.

For Gym-MuJoCo, we use the IsaacLab on-policy configurations from FlowRL as the starting point for network architectures and algorithmic hyperparameters, since the Gym-MuJoCo locomotion tasks are structurally similar to the corresponding IsaacLab locomotion tasks.
This includes policy and value network architectures, optimizer settings, rollout/update settings, and generative-policy parameters whenever applicable.
The main systematic difference is the total number of environment interactions: FlowRL's IsaacLab setting uses $100$M environment steps with massively parallel GPU simulation, whereas our Gym-MuJoCo experiments use $40$M steps because Gym-MuJoCo does not support the same scale of GPU-parallel environment interaction.

\begin{table}[t]
\centering
\caption{Hyperparameters for Playground experiments. A single configuration is used for all Playground tasks within each method.}
\label{tab:playground_hypers}
\resizebox{\linewidth}{!}{
\begin{tabular}{lccccc}
\toprule
\textbf{Hyperparameter} & \textbf{PPO} & \textbf{DPPO} & \textbf{FPO} & \textbf{GenPO} & \textbf{GSB-MDPO} \\
\midrule
\multicolumn{6}{c}{\textit{Benchmark and rollout setup}} \\
\midrule
Total environment steps & 60M & 60M & 60M & 60M & 60M \\
Number of environments & 2048 & 2048 & 2048 & 2048 & 2048 \\
Episode length & 1000 & 1000 & 1000 & 1000 & 1000 \\
Number of evaluations & 10 & 10 & 10 & 10 & 10 \\
Unroll length & 30 & 30 & 30 & 30 & 30 \\
Batch size & 1024 & 1024 & 1024 & 1024 & 1024 \\
Number of minibatches & 32 & 32 & 32 & 32 & 32 \\
Updates per batch & 16 & 16 & 16 & 16 & 16 \\
Discount factor $\gamma$ & 0.995$^\dagger$ & 0.995$^\dagger$ & 0.995$^\dagger$ & 0.995$^\dagger$ & 0.995$^\dagger$ \\
GAE $\lambda$ & 0.95 & 0.95 & 0.95 & 0.95 & 0.95 \\
Reward scaling & 10.0 & 10.0 & 10.0 & 10.0 & 10.0 \\
Normalize observations & Yes & Yes & Yes & Yes & Yes \\
Normalize advantages & Yes & Yes & Yes & Yes & Yes \\
Optimizer & Adam & Adam & Adam & Adam & Adam \\
\midrule
\multicolumn{6}{c}{\textit{Actor configuration}} \\
\midrule
Actor network & MLP, $4\times32$ & MLP, $4\times32$ & MLP, $4\times32$ & MLP, $3\times64$ & MLP, $4\times32$ \\
Actor learning rate & $3\times10^{-4}$ & $3\times10^{-4}$ & $3\times10^{-4}$ & $3\times10^{-4}$ & $1\times10^{-3}$ \\
Entropy coefficient & 0 & 0 & 0 & 0 & 0 \\
PPO clip $\epsilon$ & 0.1 & 0.2 & 0.05 & 0.2 & -- \\
\midrule
\multicolumn{6}{c}{\textit{Critic configuration}} \\
\midrule
Critic network & MLP, $5\times256$ & MLP, $5\times256$ & MLP, $5\times256$ & MLP, $5\times256$ & MLP, $5\times256$ \\
Critic learning rate & $3\times10^{-4}$ & $3\times10^{-4}$ & $3\times10^{-4}$ & $3\times10^{-4}$ & $1\times10^{-3}$ \\
Value loss coefficient & 0.25 & 0.25 & 0.25 & 1.0 & 0.1 \\
\midrule
\multicolumn{6}{c}{\textit{Generative-policy configuration}} \\
\midrule
Generation steps & -- & 10 & 10 & 5 & 8 \\
Time embedding dimension & -- & 8 & 8 & 32 & 8 \\
Policy output scale & -- & 0.25 & 0.25 & -- & 0.25 \\
Sigma schedule & -- & Constant & -- & -- & Exp. \\
$\sigma_{\max}$ / $\sigma_{\min}$ & -- & 0.05 / 0.05 & -- & -- & 3.0 / 1.0 \\
\midrule
\multicolumn{6}{c}{\textit{FPO-specific configuration}} \\
\midrule
Samples per action & -- & -- & 8 & -- & -- \\
\midrule
\multicolumn{6}{c}{\textit{GenPO-specific configuration}} \\
\midrule
Flow mix parameter & -- & -- & -- & 0.9 & -- \\
Compress coefficient & -- & -- & -- & 0.01 & -- \\
\midrule
\multicolumn{6}{c}{\textit{GSB-MDPO-specific configuration}} \\
\midrule
Step log-ratio clip & -- & -- & -- & -- & 0.1 \\
Path log-ratio clip & -- & -- & -- & -- & 0.3 \\
KL penalty coefficient & -- & -- & -- & -- & 0.8 \\
Reference mix $\eta$ & -- & -- & -- & -- & 0.0 \\
\bottomrule
\end{tabular}
}
\vspace{0.5em}
\begin{flushleft}
\footnotesize
$^\dagger$ Following the FPO Playground setup, $\gamma=0.995$ is used for most tasks; BallInCup and FingerSpin use $\gamma=0.95$.
\end{flushleft}
\end{table}

\subsection{Hyperparameters}
\label{app:hyperparameters}

For each benchmark, we use a single benchmark-level hyperparameter configuration for each method across all tasks, without task-specific tuning.
This protocol is intended to make the comparison depend on the algorithmic design rather than per-environment hyperparameter search.
In the Gym-MuJoCo benchmark, GSB-MDPO uses a cosine learning-rate schedule for the policy/flow optimizer, decaying from the reported initial learning rate to zero over training.

\paragraph{Playground.}
Table~\ref{tab:playground_hypers} summarizes the hyperparameters used for the Playground experiments.
The PPO, FPO, and DPPO configurations follow the FPO Playground setup~\citep{mcallister2025fpo}; DPPO is implemented through the denoising-MDP mode of the FPO training pipeline.
GenPO and GSB-MDPO are implemented in the same Playground training interface for fair comparison.

\paragraph{Gym-MuJoCo.}
Table~\ref{tab:gym_hypers} summarizes the hyperparameters used for the Gym-MuJoCo experiments.
For each method, we use a single hyperparameter configuration across all six Gym-MuJoCo tasks, without task-specific tuning.
The Gym-MuJoCo configurations are adapted from the FlowRL-style on-policy IsaacLab configurations, with an additional on-policy Gym-MuJoCo launcher implemented for our experiments.

\begin{table}[t]
\centering
\caption{Hyperparameters for Gym-MuJoCo experiments. A single configuration is used for all Gym-MuJoCo tasks within each method.}
\label{tab:gym_hypers}
\resizebox{\linewidth}{!}{
\begin{tabular}{lccccc}
\toprule
\textbf{Hyperparameter} & \textbf{PPO} & \textbf{DPPO} & \textbf{FPO} & \textbf{GenPO} & \textbf{GSB-MDPO} \\
\midrule
\multicolumn{6}{c}{\textit{Benchmark and rollout setup}} \\
\midrule
Total environment steps & 40M & 40M & 40M & 40M & 40M \\
Evaluation interval & 1M & 1M & 1M & 1M & 1M \\
Evaluation episodes & 10 & 10 & 10 & 10 & 10 \\
Number of environments & 1024 & 1024 & 1024 & 1024 & 1024 \\
Rollout length & 24 & 24 & 24 & 24 & 24 \\
Batch size & 6144 & 6144 & 6144 & 6144$^\ddagger$ & 6144 \\
Number of minibatches & 4 & 4 & 4 & 4$^\ddagger$ & 4 \\
Training epochs & 4 & 4 & 4 & 4 & 4 \\
Discount factor $\gamma$ & 0.99 & 0.99 & 0.99 & 0.99 & 0.99 \\
GAE $\lambda$ & 0.95 & 0.95 & 0.95 & 0.95 & 0.95 \\
Reward scaling & 1.0 & 1.0 & 1.0 & 1.0 & 1.0 \\
Normalize observations & Yes & Yes & Yes & Yes & Yes \\
Normalize advantages & Yes & Yes & Yes & Yes & Yes \\
Gradient clip norm & 1.0 & 1.0 & 1.0 & 1.0 & 1.0 \\
\midrule
\multicolumn{6}{c}{\textit{Actor configuration}} \\
\midrule
Actor network & MLP, $3\times256$ & MLP, $3\times256$ & MLP, $3\times256$ & MLP, $3\times256$ & MLP, $3\times256$ \\
Actor activation & ELU & SiLU & SiLU & SiLU & SiLU \\
Actor learning rate & $1\times10^{-4}$ & $1\times10^{-3}$ & $1\times10^{-4}$ & $1\times10^{-3}$ & $7.5\times10^{-4}$ \\
PPO clip $\epsilon$ & 0.2 & 0.2 & 0.05 & 0.2 & -- \\
Entropy coefficient & 0 & -- & -- & 0.0 & -- \\
\midrule
\multicolumn{6}{c}{\textit{Critic configuration}} \\
\midrule
Critic network & MLP, $3\times256$ & MLP, $3\times256$ & MLP, $3\times256$ & MLP, $3\times256$ & MLP, $3\times256$ \\
Critic activation & ELU & ELU & ELU & ELU & ELU \\
Critic learning rate & $1\times10^{-3}$ & $1\times10^{-3}$ & $1\times10^{-3}$ & $1\times10^{-3}$ & $1\times10^{-3}$ \\
\midrule
\multicolumn{6}{c}{\textit{Generative-policy configuration}} \\
\midrule
Generation steps & -- & 10 & 10 & 16 & 16 \\
Flow / diffusion hidden dims & -- & $3\times256$ & $3\times256$ & $3\times256$ & $3\times256$ \\
Time embedding dimension & -- & 32 & 16 & 16 & 16 \\
Flow / diffusion output scale & -- & -- & 0.25 & 0.25 & 0.25 \\
Noise schedule & -- & Cosine DDPM & -- & -- & Linear \\
Min log-prob std. & -- & 0.1 & -- & -- & -- \\
$\sigma_{\max}$ / $\sigma_{\min}$ & -- & -- & -- & -- & 3.0 / 0.3 \\
\midrule
\multicolumn{6}{c}{\textit{FPO-specific configuration}} \\
\midrule
Monte Carlo samples & -- & -- & 8 & -- & -- \\
Additive noise & -- & -- & 0.1 & -- & -- \\
\midrule
\multicolumn{6}{c}{\textit{GenPO-specific configuration}} \\
\midrule
Flow mix parameter & -- & -- & -- & 0.9 & -- \\
Compress coefficient & -- & -- & -- & 0.01 & -- \\
\midrule
\multicolumn{6}{c}{\textit{GSB-MDPO-specific configuration}} \\
\midrule
Step log-ratio clip & -- & -- & -- & -- & 0.1 \\
Path log-ratio clip & -- & -- & -- & -- & 0.4 \\
KL penalty coefficient & -- & -- & -- & -- & 0.08 \\
Reference mix $\eta$ & -- & -- & -- & -- & 0.02 \\
\bottomrule
\end{tabular}
}
\vspace{0.5em}
\begin{flushleft}
\footnotesize
$^\ddagger$ For GenPO on Humanoid-v5 only, we use batch size $4096$ and $6$ minibatches due to GPU memory constraints. All other GenPO Gym-MuJoCo tasks use the default batch size $6144$ and $4$ minibatches.
\end{flushleft}
\end{table}

\clearpage

\section{Additional Experimental Results}
\subsection{Toy Analysis of the Optimal Path-Space Update}
\label{app:toy_analysis}

We include a controlled two-dimensional toy example to visualize the closed-form optimizer in Proposition~\ref{prop:optimal_gsb_mdpo}.
The old policy $p_{\mathrm{old}}$ is a four-mode Gaussian mixture over the terminal action $x=(x_1,x_2)$.
We define a quadrant-wise preference $A(x)$ and construct the exponential-tilted target
\begin{equation}
    p^*(x)
    \propto
    p_{\mathrm{old}}(x)
    \exp\!\left(\frac{A(x)}{\beta}\right),
    \label{eq:toy_exp_tilt}
\end{equation}
which is the terminal marginal implied by the optimal path-space mirror-descent update.
The preference assigns the largest weight to quadrant QIII and smaller weights to the other quadrants, so the optimal update should increase probability mass in QIII while preserving the local mode structure inherited from the old policy.

\begin{figure}[H]
    \centering
    \includegraphics[width=\linewidth]{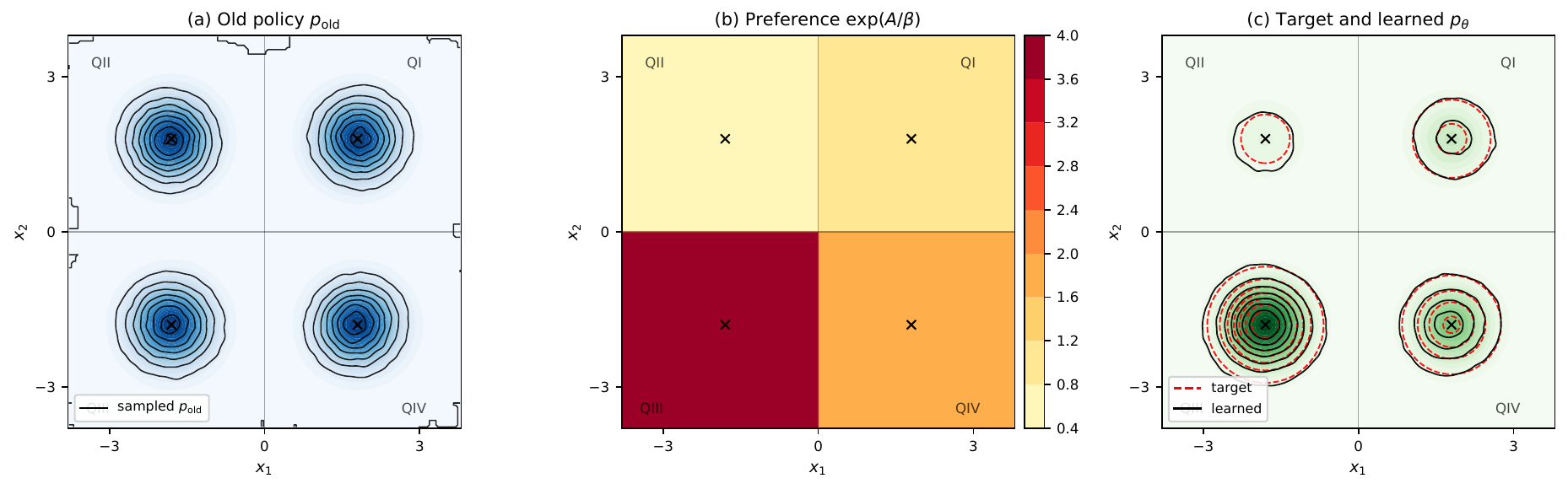}
    \caption{
    Toy visualization of the optimal GSB-MDPO update.
    Left: old policy $p_{\mathrm{old}}$ with four approximately balanced modes.
    Middle: quadrant-wise preference weights $\exp(A/\beta)$.
    Right: target exponential-tilted marginal $p^*$ and the learned distribution $p_\theta$.
    The learned policy closely matches the theoretical tilted target, increasing mass in the preferred quadrant while preserving the old policy's mode structure.
    }
    \label{fig:toy_analysis}
\end{figure}

Figure~\ref{fig:toy_analysis} confirms the behavior predicted by Proposition~\ref{prop:optimal_gsb_mdpo}.
The exponential tilting changes the relative mass of the modes according to the terminal preference while leaving the conditional path structure unchanged in the ideal solution.
In the learned distribution, the quadrant masses closely match the target distribution, with a quadrant-mass $\ell_1$ error of $0.100$.
This illustrates that GSB-MDPO implements the intended mirror-descent update: it does not simply collapse to the highest-reward mode, but reweights the old generative policy according to the advantage-induced exponential tilt.

\subsection{Step Log-Ratio Clipping Analysis}
\label{app:clip_diagnostics}

\begin{figure}[t]
    \centering
    \includegraphics[width=0.6\linewidth]{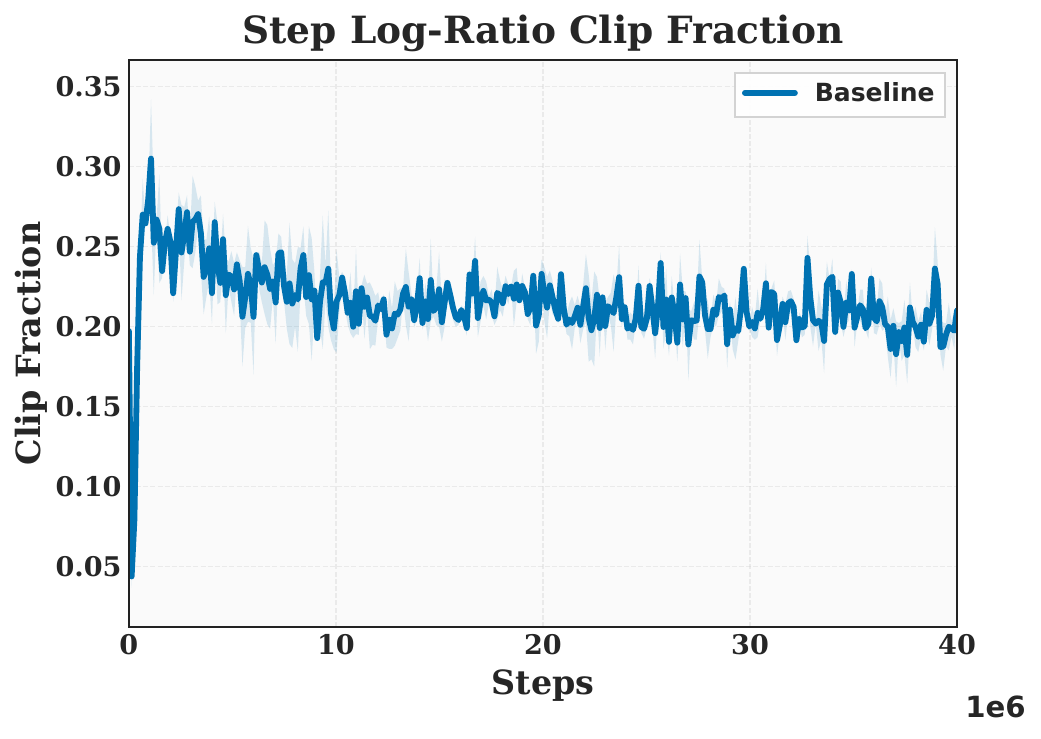}
    \caption{
    Fraction of rollout samples whose step log-ratio is clipped during GSB-MDPO training on HalfCheetah-v5.
    Results are averaged over two random seeds, with shaded regions indicating variability across seeds.
    }
    \label{fig:step_clip_ratio}
\end{figure}

To better understand the role of step-level clipping, we record the fraction of rollout samples whose step log-ratio exceeds the clipping threshold during training.
Figure~\ref{fig:step_clip_ratio} reports this diagnostic on HalfCheetah-v5 for the full GSB-MDPO baseline, averaged over two random seeds.

The clipping fraction rises quickly at the beginning of training and then stabilizes at roughly $0.2$ throughout most of optimization.
This behavior suggests that step-level clipping is neither a rarely activated corner case nor a mechanism that completely dominates learning.
Instead, it acts as a persistent stabilizer for the importance-sampled path-ratio estimator, especially during periods when the current policy moves away from the old one.

The diagnostic also helps clarify the role of clipping in GSB-MDPO.
The path-KL drift regularizer is part of the mirror-descent objective itself, whereas step log-ratio clipping is an implementation device for controlling the variance of the accumulated path likelihood ratio.
The moderate and stable clipping frequency observed in Figure~\ref{fig:step_clip_ratio} indicates that this mechanism is active often enough to matter in practice, while still leaving most stepwise likelihood ratios within the unclipped regime.

\subsection{Computational Efficiency}
\label{app:compute_efficiency}

We report representative computational cost on HalfCheetah-v5 in Table~\ref{tab:compute_efficiency}.
All methods are measured with seed $0$ for $5$M environment interactions using the same hardware configuration as Appendix~\ref{app:hardware}.
The reported wall-clock time is the end-to-end training time, and GPU memory is measured from the process tree during training.
This experiment is intended as a lightweight efficiency comparison rather than a full benchmark-wide profiling study.

\begin{table}[H]
\centering
\caption{
Representative computational cost on HalfCheetah-v5 for $5$M environment interactions.
Wall-clock time is end-to-end training time. GPU memory is measured from the training process tree.
}
\label{tab:compute_efficiency}
\begin{tabular}{lcccc}
\toprule
\textbf{Method} &
\textbf{Wall-clock} &
\textbf{Rel. time} &
\textbf{Peak GPU memory} &
\textbf{Mean GPU memory} \\
\midrule
PPO       & 6.24 min  & $1.00\times$ & 0.60 GB & 0.59 GB \\
FPO       & 6.96 min  & $1.12\times$ & 1.00 GB & 0.97 GB \\
DPPO      & 8.67 min  & $1.39\times$ & 2.50 GB & 2.38 GB \\
GenPO     & 28.73 min & $4.61\times$ & 64.98 GB & 64.13 GB \\
GSB-MDPO  & 7.39 min  & $1.19\times$ & 2.59 GB & 2.51 GB \\
\bottomrule
\end{tabular}
\end{table}

GSB-MDPO adds moderate overhead relative to Gaussian PPO due to storing denoising paths and recomputing current path likelihoods and drifts during policy updates.
However, its end-to-end runtime remains close to FPO and lower than DPPO in this representative setting.
Its GPU memory footprint is comparable to DPPO and substantially lower than GenPO, whose implementation requires much larger memory in this HalfCheetah-v5 run.
These results suggest that the path-space objective can be implemented with practical overhead relative to other multi-step generative-policy baselines.

%%%%%%%%%%%%%%%%%%%%%%%%%%%%%%%%%%%%%%%%%%%%%%%%%%%%%%%%%%%%
% NeurIPS checklist
%%%%%%%%%%%%%%%%%%%%%%%%%%%%%%%%%%%%%%%%%%%%%%%%%%%%%%%%%%%%

% \newpage
% \input{checklist.tex}

\end{document}